\documentclass[lettersize,journal]{IEEEtran}
\usepackage{amsmath,amsfonts}
\usepackage{amssymb}
\usepackage{multirow}
\usepackage[ruled,vlined]{algorithm2e}
\DontPrintSemicolon
\usepackage{array}
\usepackage{newtxtext,newtxmath}
\usepackage[caption=false,font=normalsize,labelfont=rm,textfont=rm]{subfig}
\usepackage{textcomp}
\usepackage{stfloats}
\usepackage{url}
\usepackage{verbatim}
\usepackage{graphicx}
\usepackage{cite}
\usepackage{booktabs}
\usepackage{siunitx}
\begin{document}

\title{Accelerating Human-Aware Robot Trajectory Generation via Diffusion and Consistency Distillation}

\author{Byeong-Il Ham, Hyun-Bin Kim, and Kyung-Soo Kim*
}

\maketitle

\begin{abstract}
This research proposes a constrained motion planning framework for robot manipulators in human–robot interaction (HRI). For a non-redundant manipulator with a fully specified end-effector pose, additional requirements such as collision avoidance and self-collision avoidance are difficult to handle as simple null-space secondary tasks. This limitation makes it challenging to generate feasible joint-space trajectories in HRI environments where safety and kinematic constraints must be considered simultaneously. To address this limitation, collision- and self-collision-aware trajectories are generated using Rapidly-exploring Random Tree (RRT) and RRT* algorithms, and the resulting dataset is used to train a diffusion model that generates constraint-satisfying trajectories through guided sampling. To reduce the inference time required for iterative diffusion sampling, consistency distillation is applied, and a joint-weighted jerk regularization term is incorporated into the loss function to promote smoother trajectories by penalizing abrupt changes in joint acceleration. Simulation results show that the consistency model generates 150 trajectory candidates in less than 100 ms, maintains a high episode success rate, and substantially reduces joint and end-effector jerk when jerk regularization is applied.
\end{abstract}


\section{Introduction}
\IEEEPARstart{A}{s} robot manipulators are increasingly deployed in shared workspaces with humans, human–robot collision avoidance has become an important safety requirement in human–robot interaction (HRI) systems~\cite{kwon2025channel}. In close-proximity physical HRI, robot motion should be planned to satisfy safety and feasibility requirements while ensuring that the generated trajectories are physically executable and acceptable to human users~\cite{pupa2023dynamic}. Therefore, robot motion planning should simultaneously account for the position and orientation of the end-effector (EE) as well as additional requirements such as collision and singularity avoidance. A larger number of extra tasks can increase the required degree of freedom (DoF) for simultaneous task execution. Kinematic redundancy allows secondary tasks to be performed using null-space motion. However, this approach becomes constrained when redundancy is not available~\cite{jung2023operational, dietrich2015overview}.

Conventional motion planning approaches, including graph-search, sampling-based, and optimization-based methods, have been widely used to generate feasible robot trajectories. Graph-search methods discretize the configuration space into a graph and search for feasible paths. Sampling-based methods explore the configuration space by generating and connecting samples, enabling feasible path planning in high-dimensional spaces without discretizing the entire space. Optimization-based methods formulate trajectory generation as an optimization problem with cost functions and constraints, enabling the generation of trajectories that satisfy various constraints. However, these methods may require increased computation time depending on graph resolution and constraint formulation, and incorporating all required constraints can remain challenging.

To address these limitations, this study proposes a human-aware constrained motion planning framework for robot manipulators in HRI. The proposed framework employs a diffusion model conditioned on the initial robot state, desired EE goal pose, and surrounding obstacles to generate executable joint-space trajectory candidates under human-aware constraints. The diffusion model is trained using constraint-satisfying trajectories generated by sampling-based planners, enabling it to capture trajectory patterns related to goal reaching, collision avoidance, and kinematic feasibility. In addition, jerk-aware regularization is introduced to improve trajectory smoothness, while constraint-guided sampling is employed to enhance the feasibility of the generated trajectory candidates. To reduce the inference time required for iterative diffusion sampling, consistency distillation is applied to derive a fast trajectory generator from the diffusion model. The main contributions of this study are summarized as follows:

\begin{itemize}
    \item A constrained motion planning formulation is presented to address secondary-task requirements under limited kinematic redundancy.
    \item A consistency-distilled trajectory generator is developed to reduce the inference time of iterative diffusion sampling while maintaining trajectory feasibility.
    \item A joint-weighted jerk regularization term is introduced to improve trajectory smoothness by penalizing rapid changes in joint acceleration across individual joints.
\end{itemize}

\subsection{Related Works}
In robot manipulator path planning, sampling-based motion planning has been widely used to search for feasible paths in the configuration space. Rapidly-exploring Random Tree (RRT) and its optimal variant, RRT*, are widely used sampling-based planners that expand search trees in the configuration space to find collision-free paths from an initial configuration to a goal configuration or goal region~\cite{lavalle1998rapidly,karaman2011sampling}. However, RRT-based planners may exhibit degraded planning performance in complex environments, such as narrow passages, and various RRT variants have been proposed to address this issue\cite{feng2024adaptive,qi2023path}.

Jiang \textit{et al.}~\cite{jiang2022path} proposed an improved RRT method for manipulator path planning in complex multi-obstacle environments by combining hybrid constrained sampling, artificial potential field-based expansion, adaptive step size, pruning, and local replanning. However, despite its improved planning efficiency, the method remains an iterative sampling-based planner based on tree expansion, collision checking, and replanning, which can be computationally demanding for repeated trajectory generation in HRI scenarios. Shen \textit{et al.}~\cite{shen2023adaptive} presented an adaptive manipulability-based RRT* planner that incorporates path length and manipulability into the planning cost and adjusts the step size based on manipulability and goal distance. However, because the planner is implemented offline and produces inherently nonsmooth paths, additional smoothing and acceleration strategies may be required for fast trajectory generation. These studies show that sampling-based planners can address multiple planning constraints, but reducing computational cost remains challenging.

Optimization-based planning is another widely used approach for robot trajectory generation~\cite{schulman2013finding,zucker2013chomp,kalakrishnan2011stomp}. It formulates trajectory generation as an optimization problem over costs and constraints, allowing requirements such as obstacle avoidance and smoothness to be encoded in the planning objective or constraints.

Marcucci \textit{et al.}~\cite{marcucci2023motion} presented a graph-of-convex-sets-based motion planning framework that formulates collision-free trajectory generation as a convex optimization problem over safe convex regions around obstacles. Nevertheless, its guarantees depend on the convex decomposition of free space and the selected trajectory parameterization.
Wen and Pagilla~\cite{wen2022path} developed a path-constrained, collision-free optimal trajectory planning method that accounts for waypoint tracking, obstacle and self-collision avoidance, kinematic and dynamic constraints, and time-jerk optimality. Despite these advantages, the method assumes predefined waypoints and static obstacles, and its nonlinear programming formulation can become computationally demanding as the number of waypoints and grid points increases or additional nonlinear constraints are imposed. Ji \textit{et al.}~\cite{ji2023convex} developed a B-spline-based convex optimization method for time-optimal trajectory planning with hard jerk constraints for manipulators. However, the method is limited to trajectory optimization along a predefined path, and its convex restriction of jerk constraints sacrifices some time optimality for improved smoothness and computational efficiency.

Recently, diffusion models have gained increasing attention in robot motion planning because they can model trajectory-level distributions and generate diverse trajectory candidates~\cite{janner2022planning,huang2023diffusion,huang2024diffusionseeder}. Carvalho \textit{et al.}~\cite{carvalho2023motion} proposed motion planning diffusion, which learns a diffusion-based trajectory prior from collision-free trajectories and performs cost-guided posterior sampling for start-to-goal motion planning under collision-avoidance and joint-limit constraints. Ma \textit{et al.}~\cite{ma2024hierarchical} presented hierarchical diffusion policy, which combines a language-guided next-best-pose agent with a goal-conditioned robot kinematics diffuser for kinematics-aware joint-position trajectory generation in multi-task robotic manipulation. Diffusion-based robot motion generation often requires multiple iterative denoising steps, which can lead to long inference time~\cite{prasad2024consistency}. To address this issue, Lu \textit{et al.}~\cite{lu2024manicm} developed ManiCM, a real-time 3D diffusion policy that imposes consistency constraints on the point-cloud-conditioned diffusion process to enable one-step robotic action generation.

\subsection{Overview}
The remainder of this paper is organized as follows. Section~\ref{sdiffusion} formulates the human-aware constrained trajectory planning problem and presents the diffusion-based trajectory generation framework. Section~\ref{sconsistency} describes the consistency distillation method for reducing the inference time of iterative diffusion sampling. Section~\ref{svalidations} presents the simulation setup and validation results. Finally, Section~\ref{sdiscussionandconclusion} concludes the paper.

\section{Human-Aware Trajectory Planning with Diffusion Models}
\label{sdiffusion}
This section presents the proposed human-aware joint-space trajectory planning framework based on a conditional diffusion model. We first formulate the constrained planning problem and define the trajectory dataset. We then describe how the diffusion model learns the conditional trajectory distribution and generates trajectory candidates through constraint-guided sampling.

\subsection{Constrained Planning Formulation and Trajectory Dataset}
This study addresses joint-space trajectory generation for a robot manipulator whose EE is required to reach a target near the human hand while maintaining safety. In HRI, trajectory planning involves not only the desired EE goal pose but also additional safety and feasibility requirements, such as collision avoidance, self-collision avoidance, and manipulability. The problem is formulated as finding a constrained joint-space trajectory toward an inverse-kinematics (IK) goal selected to maximize manipulability at the target. The trajectory is required to avoid collisions with the human arm, torso, and surrounding obstacles, as well as self-collisions.

Let $\mathbf{q}_t \in \mathbb{R}^n$ denote the joint vector of an $n$-DoF manipulator at waypoint $t$, where $t=0,1,\cdots,T-1$ and $T$ is the number of waypoints. The joint-space trajectory is expressed as follows:
\begin{equation}
\label{trajectoryData}
    \mathbf{q}_{0:T-1} = [\mathbf{q}_0, \mathbf{q}_1, \ldots, \mathbf{q}_{T-1}] \in \mathbb{R}^{T \times n}.
\end{equation}
The waypoint $t=0$ represents the initial joint configuration of the manipulator, whereas $t=T-1$ represents the final joint configuration. The desired EE goal pose is defined by the goal position $\mathbf{x}_g \in \mathbb{R}^3$ and goal orientation $\mathbf{R}_g \in SO(3)$. In this study, we consider a 6-DoF manipulator, i.e., $n=6$, for which a fully specified EE pose leaves limited kinematic redundancy. Therefore, the desired EE goal pose is converted into joint-space goal candidates through numerical IK, and the candidate with the highest manipulability is selected as the goal configuration. For numerical IK, $\xi_a^l$ denotes the joint estimate at iteration $l$ of the $a$-th IK attempt. At each iteration, the solver computes a 6D pose error composed of the position error and the logarithmic orientation error as follows:
\begin{equation}
\label{errorIK}
    \mathbf{e}_{a}^{l}
    =
    \begin{bmatrix}
        \mathbf{x}_g-\mathbf{x}_{ee}\left(\boldsymbol{\xi}_{a}^{l}\right)\\
        \log(\mathbf{R}_g
        \mathbf{R}_{ee}\left(\boldsymbol{\xi}_{a}^{l}\right)^{\top})^{\vee}
    \end{bmatrix}
\end{equation}
here $\mathbf{x}_{ee}$ and $\mathbf{R}_{ee}$ denote the EE position and orientation obtained from the current $\xi$ , respectively, and $\log(\cdot)^{\vee}$ maps the rotation error matrix to its vector representation. The joint estimate is updated using a damped least-squares step:

\begin{equation}
    \Delta \boldsymbol{\xi}_{a}^{l}
    =
    \left(
    \mathbf{J}_{a}^{l\top}
    \mathbf{J}_{a}^{l}
    +
    \lambda \mathbf{I}
    \right)^{-1}
    \mathbf{J}_{a}^{l\top}
    \mathbf{e}_{a}^{l}
\end{equation}
where $\mathbf{J}_{a}^{l}$ denotes the EE Jacobian evaluated at $\boldsymbol{\xi}_a^l$, and $\lambda$ is a damping coefficient. An IK candidate is accepted when the position and orientation errors in \eqref{errorIK} fall below predefined thresholds. After each update, the joint estimate is normalized to the range $[-\pi,\pi]$. Among the IK candidates that satisfy collision- and self-collision-free conditions, the candidate with the highest manipulability is selected. The manipulability measure is defined using the Jacobian $\mathbf{J}$ at $\boldsymbol{\xi}$ as
\begin{equation}
\label{manipulability}
    m(\boldsymbol{\xi}) = \sqrt{\det\left(\mathbf{J}(\boldsymbol{\xi})\mathbf{J}(\boldsymbol{\xi})^\top\right)}.
\end{equation}
Let $\mathbf{q}_g^{*}$ denote the IK candidate that maximizes $m(\boldsymbol{\xi})$. The goal configuration is then defined as $\mathbf{q}_{T-1}=\mathbf{q}_g^{*}$.

Given the initial and goal joint configurations, RRT and RRT* generate the intermediate waypoints of the joint-space trajectory. During planning, each candidate edge is validated by interpolating configurations along the segment and checking joint limits, environment collisions, and self-collisions. The resulting collision-free path is then resampled into a fixed-length trajectory with $T$ waypoints, and the resampled trajectory is checked again for joint-limit, collision, and self-collision constraints. Each dataset sample consists of the initial joint configuration, desired EE goal pose, obstacle representation, and the fixed-length joint-space trajectory generated by RRT or RRT*. The resulting dataset provides training data for the conditional diffusion model, allowing it to learn the distribution of feasible joint-space trajectories under human-aware task constraints.

\begin{figure*}[t!]
    \centering
        \includegraphics[width=2.1\columnwidth]{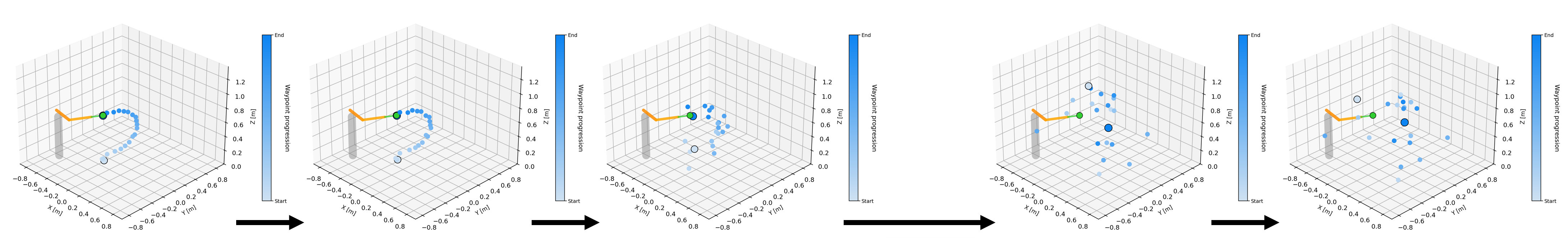}
        \caption{
        Process of noise injection. $k=1,2,25,150,199$, shown from left to right. Gray represents torso, orange
indicates the human arm, green denotes the goal pose, and blue denotes EE waypoints.
        }    
    \label{noiseInjection}
\end{figure*}

\subsection{Conditional Diffusion Model}
\label{trainDiffusion}

The conditional diffusion model learns a denoising prior over fixed-length joint-space trajectories. The model condition consists of the initial joint configuration, desired EE goal pose, and obstacle representation. The desired EE goal pose is encoded as a 9-dimensional vector, consisting of the 3D goal position and the first two column vectors of the goal orientation matrix. The obstacle representation is given as a set of capsule features, with each capsule represented by two endpoints and a radius:

\begin{equation}
\label{obstacles}
    \mathbf{o}_i
    =
    \left[
    \mathbf{p}_{i,1}^{\top},
    \mathbf{p}_{i,2}^{\top},
    \rho_i
    \right]^{\top}
    \in \mathbb{R}^{7}
\end{equation}
where $\mathbf{p}_{i,1},\mathbf{p}_{i,2} \in \mathbb{R}^3$ are the capsule endpoints and $\rho_i$ is the capsule radius. The obstacle representation is encoded using a PointNet-style set encoder. Each 7D capsule feature is independently mapped to a 64-dimensional feature through a shared two-layer multi-layer perceptron (MLP) with Mish activation, followed by symmetric max pooling over the obstacle dimension to obtain a fixed-dimensional global obstacle feature.

These conditioning features are concatenated and projected by a condition MLP. The resulting condition embedding is added to the sinusoidal diffusion timestep embedding and injected into all residual temporal blocks, which are residual 1D convolutional blocks applied along the waypoint dimension. The backbone is implemented as a temporal U-Net~\cite{janner2022planning}, where the noised joint trajectory is projected into temporal feature channels, processed through residual temporal blocks with downsampling and upsampling, and mapped back to the trajectory dimension to predict the injected noise.

During training, the forward diffusion process corrupts the clean trajectory by adding Gaussian noise $\boldsymbol{\epsilon}$
\begin{equation}
\label{noiseInjectionEqu}
    \mathbf{q}^{k}_{\mathrm{d}}
    =
    \sqrt{\bar{\alpha}_{k}}\mathbf{q}_{0:T-1}
    +
    \sqrt{1-\bar{\alpha}_{k}}\boldsymbol{\epsilon},
    \qquad
    \boldsymbol{\epsilon}\sim\mathcal{N}(\mathbf{0},\mathbf{I}).
\end{equation}
The diffusion timestep $k \in K$, distinct from the trajectory waypoint index $t$, represents the noise level. The parameter $\alpha_k$ denotes the noise schedule, $\beta_k = 1 - \alpha_k$ , and $\bar{\alpha}_k$ denotes its cumulative product. 
Fig.~\ref{noiseInjection} visualizes the EE trajectory computed from $\mathbf{q}^k_\mathrm{d}$ via forward kinematics at different diffusion timesteps. In the blue trajectory, lower saturation indicates waypoints closer to $t=0$, whereas higher saturation indicates waypoints with larger $t$. As $k$ increases, the noised trajectory becomes increasingly corrupted and approaches Gaussian noise.
 
Given the noised trajectory $\mathbf{q}^{k}_{\mathrm{d}}$, diffusion timestep $k$, and task condition $\mathbf{c}$, the denoising network predicts the injected noise. The task condition $\mathbf{c}$ consists of the initial joint configuration, the desired EE goal pose, and the obstacle set $\mathcal{O}$, whose elements are defined in \eqref{obstacles}.

\begin{equation}
\label{noiseDiffusionPrediction}
    \hat{\boldsymbol{\epsilon}}
    =
    \boldsymbol{\epsilon}_{\theta}^{\mathrm{D}}
    \left(
    \mathbf{q}^{k}_{\mathrm{d}},
    k,
    \mathbf{c}
    \right)
\end{equation}
here $\boldsymbol{\epsilon}_{\theta}^{\mathrm{D}}$ is the denoising network of the diffusion model with learnable parameters $\theta$, and $\hat{\boldsymbol{\epsilon}}$ denotes the predicted noise. The training loss is defined as the mean squared error (MSE) between the injected noise $\boldsymbol{\epsilon}$ and $\hat{\boldsymbol{\epsilon}}$

\begin{equation}
\label{diffusionMSE}
    \mathcal{L}_{\mathrm{MSE}}
    =
    \mathbb{E}_{\mathbf{q}_{0:T-1}, k, \boldsymbol{\epsilon}}
    \left[
    \left\|
    \boldsymbol{\epsilon}
    -
    \boldsymbol{\epsilon}_{\theta}^{\mathrm{D}}
    \left(
    \mathbf{q}^{k}_{\mathrm{d}},
    k,
    \mathbf{c}
    \right)
    \right\|_2^2
    \right].
\end{equation}

\begin{figure*}[t!]
    \centering
        \includegraphics[width=2.1\columnwidth]{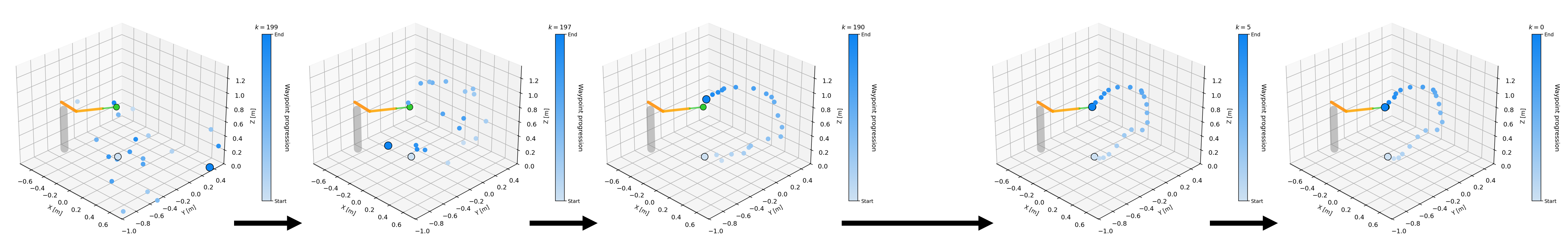}
        \caption{
        Constraint-guided diffusion inference process. Each panel shows the EE trajectory obtained by applying forward kinematics to the predicted clean trajectory estimate at the displayed diffusion timestep during guided reverse sampling. The diffusion timestep decreases from $k=199$ to $k=0$.
        }
    \label{diffusionInference}
\end{figure*}

\subsection{Constraint-Guided Sampling}

After training, the diffusion model generates trajectory candidates through the reverse denoising process. Starting from Gaussian noise, the model predicts the noise at each diffusion timestep $k$ and progressively refines the trajectory toward the learned trajectory distribution. At each reverse diffusion step, the denoising network predicts the injected noise from the current noised trajectory and task condition, and the predicted mean of the reverse transition is computed as

\begin{equation}
\label{meanDiffusion}
    \boldsymbol{\mu}_{\theta}
    =
    \frac{1}{\sqrt{\alpha_k}}
    \left(
    \mathbf{q}^{k}
    -
    \frac{\beta_k}{\sqrt{1-\bar{\alpha}_k}}
    \boldsymbol{\epsilon}_{\theta}^{\mathrm{D}}
    \left(
    \mathbf{q}^{k},
    k,
    \mathbf{c}
    \right)
    \right).
\end{equation}

In the reverse process, the next trajectory state is sampled by adding Gaussian noise to the predicted mean $\boldsymbol{\mu}_{\theta}$ of the reverse transition:
\begin{equation}
\label{reverseDiffusion}
    \mathbf{q}^{k-1}
    =
    \boldsymbol{\mu}_{\theta}
    +
    \sqrt{\beta_k}\mathbf{z},
    \qquad
    \mathbf{z}\sim\mathcal{N}(\mathbf{0},\mathbf{I}).
\end{equation}
This reverse update is repeated until $k=0$, which yields a clean trajectory candidate conditioned on $\mathbf{c}$. However, since the denoising model is trained on a finite trajectory dataset, the learned trajectory prior alone may be insufficient to avoid obstacles in configurations that are not well represented in the training data.

To improve collision avoidance during inference, the predicted mean is corrected before sampling the next trajectory state using the gradient of a differentiable constraint cost $\mathcal{C}$, which includes environment collision and self-collision costs:

\begin{algorithm}[t!]
\caption{Consistency distillation training}
\label{algorithmCDTraining}

\For{each training iteration}{
   \textbf{get} $\mathbf{q}_{0:T-1}$ in \eqref{trajectoryData} and task condition $\mathbf{c}$\;
   \textbf{get} $k$ and $n$ s.t. $0 \le k < k+n \le K$\;
   \textbf{compute} $\mathbf{q}_{\mathrm{d}}^{k+n}$ from $\mathbf{q}_{0:T-1}$ using \eqref{noiseInjectionEqu}\;
   \textbf{predict} teacher noise 
   $\boldsymbol{\epsilon}_{\Phi}^{\mathrm{D}}\left(\mathbf{q}_{\mathrm{d}}^{k+n}, k+n, \mathbf{c}\right)$\;
   \textbf{compute} teacher clean estimate \eqref{consistencyFunction} $\to \hat{\mathbf{q}}^{0,\Phi}$\;
   \textbf{compute} teacher-guided noised state \eqref{teacherGuidedState} $\to \mathbf{q}^{k,\Phi}$\;
   \textbf{compute} reference clean estimate \eqref{consistencyFunction} $\to f_{\theta^{-}}\left(\mathbf{q}^{k,\Phi}, k, \mathbf{c}\right)$\;
   \textbf{compute} student clean estimate \eqref{consistencyFunction}
   $\to f_{\theta}\left(\mathbf{q}_{\mathrm{d}}^{k+n}, k+n, \mathbf{c}\right)$\;
   \textbf{update} student network by minimizing $\mathcal{L}_{\mathrm{CD}}$ in \eqref{lossCD}\;
   \textbf{update} EMA target network parameters $\theta^{-}$\;
}
\end{algorithm}

\begin{equation}
    \tilde{\boldsymbol{\mu}}_{\theta}
    =
    \boldsymbol{\mu}_{\theta}
    -
    \eta_k
    \nabla_{\boldsymbol{\mu}_{\theta}}
    \mathcal{C}
    \left(
    \boldsymbol{\mu}_{\theta}
    \right)
\end{equation}
where \(\tilde{\boldsymbol{\mu}}_{\theta}\) denotes the guided reverse mean and \(\eta_k\) denotes the timestep-dependent guidance scale. The corrected mean is then used in place of the original reverse mean in \eqref{meanDiffusion} to sample the next trajectory state. This guidance steers the reverse denoising process toward lower environment collision and self-collision costs while keeping the generated trajectory close to the learned trajectory prior. Fig.~\ref{diffusionInference} illustrates this inference process by visualizing the EE trajectory computed from the predicted clean trajectory estimate at different diffusion timesteps. As the diffusion timestep decreases, the estimate is progressively refined into a structured EE trajectory, while the guidance correction promotes avoidance of the human body, surrounding obstacles, and robot manipulator self-collisions.

\section{Consistency Distillation for Fast Trajectory Generation}
\label{sconsistency}

Diffusion models typically require many sampling steps during the reverse denoising process, which increases inference time and limits their applicability to fast trajectory generation. To address this limitation, we apply consistency distillation to reduce the number of denoising steps required for trajectory generation, thereby decreasing inference time. To improve trajectory quality, a joint-weighted jerk regularization term is further added to the consistency training loss function, penalizing rapid changes in joint acceleration across individual joints.

\subsection{Consistency Distillation and Few-Step Sampling}
To reduce the inference time required for trajectory generation, we distill the diffusion model described in Section~\ref{trainDiffusion} into a consistency model. Unlike the diffusion model, which relies on many reverse denoising steps, the consistency model is trained to map noised trajectory states from different diffusion timesteps to a consistent clean trajectory estimate. The consistency model adopts the same temporal U-Net backbone as the diffusion model.

\begin{equation}
\label{consistencyFunction}
    \hat{\mathbf{q}}^{0,\theta}
    =
    f_{\theta}
    \left(
    \mathbf{q}_{\mathrm{d}}^{k},
    k,
    \mathbf{c}
    \right)
    =
    \frac{
    \mathbf{q}_{\mathrm{d}}^{k}
    -
    \sqrt{1-\bar{\alpha}_{k}}
    \boldsymbol{\epsilon}_{\theta}^{\mathrm{C}}
    \left(
    \mathbf{q}_{\mathrm{d}}^{k},
    k,
    \mathbf{c}
    \right)
    }
    {\sqrt{\bar{\alpha}_{k}}}
\end{equation}

\begin{equation}
\label{teacherGuidedState}
    \mathbf{q}^{k,\Phi}
    =
    \sqrt{\bar{\alpha}_{k}}
    \hat{\mathbf{q}}^{0,\Phi}
    +
    \sqrt{1-\bar{\alpha}_{k}}
    \boldsymbol{\epsilon}_{\Phi}^{\mathrm{D}}
    \left(
    \mathbf{q}_{\mathrm{d}}^{k+n},
    k+n,
    \mathbf{c}
    \right)
\end{equation}
here \(f_{\theta}\) denotes the online student consistency model with learnable parameters \(\theta\), whose output is the clean trajectory estimate corresponding to diffusion timestep 0. The term \(\boldsymbol{\epsilon}_{\theta}^{\mathrm{C}}\) denotes the noise-prediction network of the consistency model. Let $\Phi$ denote the frozen parameter set of the pretrained diffusion model $\boldsymbol{\epsilon}_{\Phi}^{\mathrm{D}}$, which is used as the teacher network during consistency distillation. The teacher-based clean estimate \(\hat{\mathbf{q}}^{0,\Phi}\) is reconstructed from \(\mathbf{q}_{\mathrm{d}}^{k+n}\), and the teacher-guided noised state \(\mathbf{q}^{k,\Phi}\) is obtained by re-noising \(\hat{\mathbf{q}}^{0,\Phi}\) to timestep \(k\).

Algorithm~\ref{algorithmCDTraining} summarizes the consistency distillation procedure. An exponential moving average (EMA) target consistency model with parameters \(\theta^{-}\) is also maintained as a slowly updated copy of the online student. Given a clean trajectory $\mathbf{q}_{0:T-1}$, a noised trajectory is first generated at a selected teacher timestep $k+n$, where $k$ denotes the earlier diffusion timestep and $n \in \mathbb{N}$ denotes the distillation interval. The teacher network predicts the noise component of this noised trajectory, and this prediction is used to construct an intermediate noised state $\mathbf{q}^{k,\Phi}$. The EMA target network takes $\mathbf{q}^{k,\Phi}$ as input and produces a reference clean trajectory estimate. The online student network is then trained to produce the same clean estimate from the original noised trajectory $\mathbf{q}^{k+n}_\mathrm{d}$.

\begin{equation}
\label{lossCD}
    \mathcal{L}_{\mathrm{CD}}
    =
    \mathbb{E}
    \left[
    \bar{\alpha}_{k+n}
    \left\|
    f_{\theta}
    \left(
    \mathbf{q}_{\mathrm{d}}^{k+n},
    k+n,
    \mathbf{c}
    \right)
    -
    f_{\theta^{-}}
    \left(
    \mathbf{q}^{k,\Phi},
    k,
    \mathbf{c}
    \right)
    \right\|_2^2
    \right].
\end{equation}
This loss penalizes the discrepancy between the clean trajectory estimate produced by the online student from the more corrupted noised trajectory \(\mathbf{q}_{\mathrm{d}}^{k+n}\) and the reference estimate produced by the EMA target network from $\mathbf{q}^{k,\Phi}$. The EMA target parameters are updated throughout training to prevent abrupt changes in the target output, thereby stabilizing the consistency distillation process.

\begin{algorithm}[t!]
\caption{Few-step consistency sampling}
\label{algorithmCDSampling}

\textbf{get} selected timesteps $\{k_1,k_2,\ldots,k_M\}$ with $k_1=K$ and $k_M=0$\;
\textbf{initialize} $\mathbf{q}^{k_1}\sim\mathcal{N}(\mathbf{0},\mathbf{I})$\;

\For{$m \gets 1$ \KwTo $M$}{
   \textbf{compute} clean trajectory estimate 
   $\hat{\mathbf{q}}^{0,\theta} \gets f_{\theta}\left(\mathbf{q}^{k_m}, k_m, \mathbf{c}\right)$ using \eqref{consistencyFunction}\;
   \textbf{set} $\hat{\mathbf{q}}^{0,\theta}_{0} \gets \mathbf{q}_{0}$\;
   \textbf{compute} constraint guidance to $\hat{\mathbf{q}}^{0,\theta}$\;
   \textbf{set} $\hat{\mathbf{q}}^{0,\theta}_{0} \gets \mathbf{q}_{0}$\;

   \If{$m < M$}{
       \textbf{get} $\boldsymbol{\epsilon}\sim\mathcal{N}(\mathbf{0},\mathbf{I})$\;
       \textbf{compute}
       $\mathbf{q}^{k_{m+1}}
       \gets
       \sqrt{\bar{\alpha}_{k_{m+1}}}\hat{\mathbf{q}}^{0,\theta}
       +
       \sqrt{1-\bar{\alpha}_{k_{m+1}}}\boldsymbol{\epsilon}$\;
   }
}
\textbf{return} $\hat{\mathbf{q}}^{0,\theta}$\;
\end{algorithm}

\begin{figure}[t!]
    \centering
        \includegraphics[width=1.0\columnwidth]{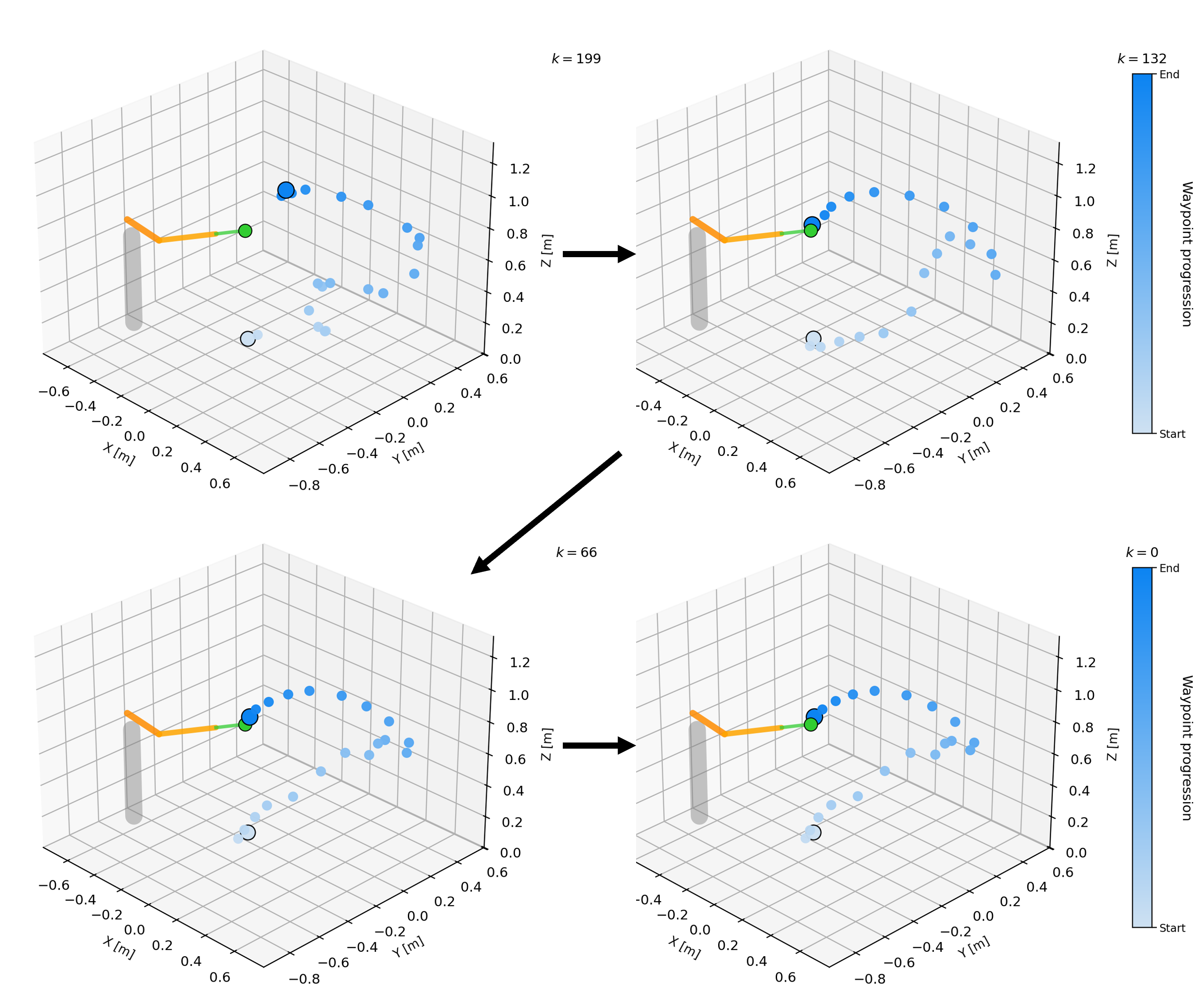}
        \caption{
        Few-step consistency sampling process. Each panel shows the EE trajectory obtained by applying forward kinematics to the guided clean trajectory estimate $\hat{\mathbf{q}}^{0,\theta}$ in Algorithm~\ref{algorithmCDSampling} at the selected timesteps $k=199,132,66,0$.
        }    
    \label{consistencyInference}
\end{figure}

During inference, the consistency model generates trajectory candidates through few-step sampling over a selected set of diffusion timesteps, as summarized in Algorithm~\ref{algorithmCDSampling}. Let $\{k_1,k_2,\ldots,k_M\}$ denote the selected timestep sequence, where $k_1=K$ is the maximum diffusion timestep and $k_M=0$. Starting from Gaussian noise $\mathbf{q}^{k_1}$, the consistency model predicts a clean trajectory estimate $\hat{\mathbf{q}}^{0,\theta}$ at each timestep $k_m$. The first waypoint of the predicted trajectory is fixed to the initial joint configuration $\mathbf{q}_0$. If $m<M$, the predicted clean trajectory estimate is re-noised to the next selected timestep $k_{m+1}$, and the resulting $\mathbf{q}^{k_{m+1}}$ is used as the input to the next consistency step. After the final timestep, the last guided clean trajectory estimate $\hat{\mathbf{q}}^{0,\theta}$ is returned as the generated trajectory. Fig.~\ref{consistencyInference} visualizes this few-step consistency sampling process by showing the EE trajectory computed from the guided clean trajectory estimate $\hat{\mathbf{q}}^{0,\theta}$ at each selected timestep.

\subsection{Joint-Weighted Jerk Regularization}
The generated joint-space trajectory must satisfy collision-free constraints while remaining sufficiently smooth for execution by the robot manipulator. Since large joint jerk can increase EE jerk, torque variation, and actuator load~\cite{zhang2021time}, a jerk penalty is incorporated into the training loss function.

The discrete joint jerk is computed from the clean trajectory estimate $\hat{\mathbf{q}}^{0,\theta}$ predicted by a model. Since the trajectory is represented by fixed-length waypoints, the joint jerk at waypoint $t$, denoted by $\Delta^3 \hat{\mathbf{q}}^{0,\theta}_{t}$, is approximated by the third-order finite difference as
\begin{equation}
    \Delta^3 \hat{\mathbf{q}}^{0,\theta}_{t}
    =
    \hat{\mathbf{q}}^{0,\theta}_{t+3}
    -
    3\hat{\mathbf{q}}^{0,\theta}_{t+2}
    +
    3\hat{\mathbf{q}}^{0,\theta}_{t+1}
    -
    \hat{\mathbf{q}}^{0,\theta}_{t}.
\end{equation}
The jerk of each joint is incorporated into the loss with three weighting factors. First, a manipulability-based weight assigns a larger penalty to low-manipulability configurations, suppressing joint variations near singular configurations. Second, a joint-wise weight based on the column norm of the Jacobian penalizes each joint in proportion to its translational sensitivity at the EE. Third, a temporal weight assigns a larger penalty to waypoints closer to the goal configuration at $T-1$, improving smoothness near the goal pose.
Based on these weighting factors, the jerk loss term is defined as follows:
\begin{equation}
\label{lossJerk}
    \mathcal{L}_{\mathrm{jerk}}
    =
    \mathbb{E}_{\hat{\mathbf{q}}^{0,\theta},t}
    \left[
    w_t^{\mathrm{time}}
    w_t^{\mathrm{manip}}
    \left\|
    \Delta^3 \hat{\mathbf{q}}^{0,\theta}_{t}
    \right\|_{\mathbf{W}_t^{\mathrm{joint}}}^{2}
    \right]
\end{equation}
here $w_t^{\mathrm{time}}$ and $w_t^{\mathrm{manip}}$ denote the temporal and manipulability-based weights, respectively. The weighted norm denotes the joint-wise weighted squared jerk and is defined as
\begin{equation}
    \left\|\mathbf{x}\right\|_{\mathbf{W}_t^{\mathrm{joint}}}^{2}
    =
    \mathbf{x}^{\top}
    \mathbf{W}_t^{\mathrm{joint}}
    \mathbf{x},
\end{equation}
where
$\mathbf{W}_t^{\mathrm{joint}} = \mathrm{diag}(w_{t,1}^{\mathrm{joint}},\ldots,w_{t,n}^{\mathrm{joint}})$
contains the joint-wise weights computed from the column norms of the Jacobian.

For diffusion model training, the MSE loss in \eqref{diffusionMSE} is extended by adding the jerk loss term in \eqref{lossJerk} as follows:
\begin{equation}
\label{diffusionTotLoss}
    \mathcal{L}_{\mathrm{diffusion}} = \mathcal{L}_{\mathrm{MSE}} + \lambda_{\mathrm{jerk}}\mathcal{L}_{\mathrm{jerk}}\,.
\end{equation}
The coefficient $\lambda_{\mathrm{jerk}}$ denotes the weight of the jerk loss term. The diffusion model trained with \eqref{diffusionTotLoss} is then distilled into the consistency model, and the consistency training loss function is also extended with the jerk loss term as follows:
\begin{equation}
\label{consistencyTotLoss}
    \mathcal{L}_{\mathrm{consistency}} = \mathcal{L}_{\mathrm{CD}} + \lambda_{\mathrm{jerk}}\mathcal{L}_{\mathrm{jerk}}\,.
\end{equation}

\section{Validation}
\label{svalidations}

\subsection{Simulation Setup}

\begin{figure}[t]
    \centering
        \includegraphics[width=0.8\columnwidth]{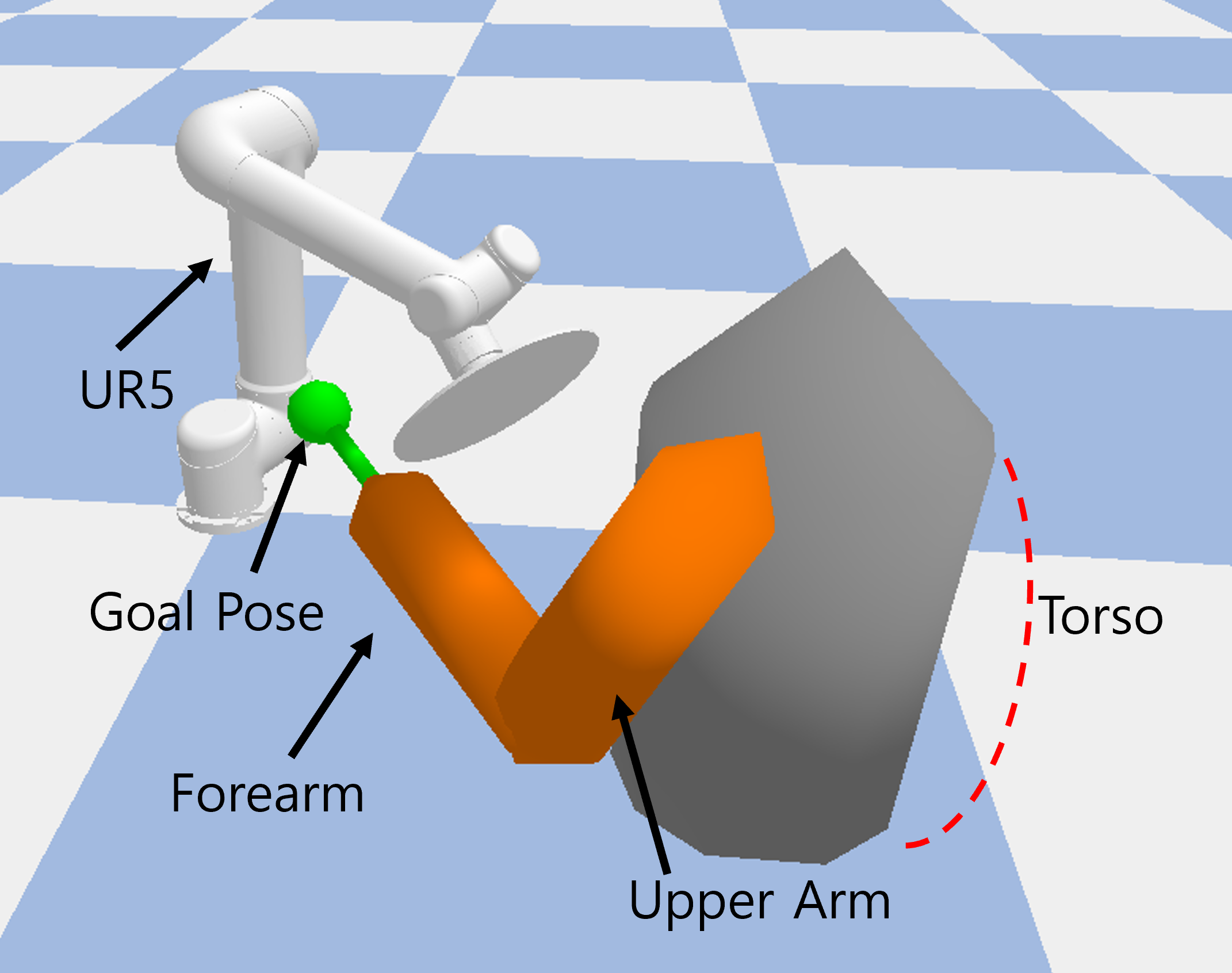}
        \caption{
        Simulation environment for human-aware trajectory generation. The UR5 manipulator reaches a goal pose near the human hand while avoiding collision primitives representing the torso, upper arm, and forearm. White denotes the UR5 manipulator, green denotes the goal pose, orange denotes the human arm, and gray denotes the torso.
        }    
    \label{simSetup}
\end{figure}

The proposed method was evaluated in simulation using a 6-DoF UR5 collaborative robot model equipped with a force/torque sensor and a contact plate at the EE. The diffusion and consistency models were trained and evaluated on a single NVIDIA GeForce RTX 5070 Ti GPU running Ubuntu 24.04. The models were implemented in PyTorch, and Pinocchio~\cite{carpentier2019pinocchio} was used for robot kinematics and dynamics.

Fig.~\ref{simSetup} shows the simulation environment used for validation. A total of 51,000 trajectory datasets were generated using RRT and RRT* with a ratio of 2:8. In each dataset, the initial joint configuration, human position, torso length, and arm length were randomly generated. To account for variations in human arm geometry, the arm length was determined by uniformly sampling one of three predefined arm types. From the generated trajectory datasets, 42,000 datasets were selected for training, 5,000 for validation, and 500 for testing.

For each test scene, each model generated 150 trajectory candidates. The generated trajectories were represented as a candidate batch \(\hat{\mathbf{q}}^{0} \in \mathbb{R}^{B \times T \times n}\), where \(B\) denotes the number of candidates. A trajectory candidate was counted as successful when the EE pose computed from the final predicted joint configuration $\hat{\mathbf{q}}_{T-1}^{0}$ reached the goal EE pose within a position error of $5~\mathrm{cm}$ and an orientation error of $10^\circ$.
 If multiple trajectory candidates satisfied the task constraints, the candidate with the highest manipulability~\eqref{manipulability} at the final configuration was selected.

\subsection{Diffusion Inference}
\label{ssDiffusionInference}

\begin{table}[t!]
\centering
\caption{Diffusion inference results with and without jerk regularization}
\resizebox{\columnwidth}{!}{
\begin{tabular}{lccc}
\hline
                         & w/ jerk & w/o jerk & Unit \\ \hline\hline
Inference time           & 5.7 & 5.7 & s \\
Success rate             & 99.60   & 98.80   & \% \\
Manipulability           & 0.0931  & 0.0930  & -- \\
Joint jerk               & $6.04{\times}10^{2}$ & $9.57{\times}10^{2}$ & $\mathrm{rad^2/s^6}$ \\
EE jerk                  & $2.04{\times}10^{2}$ & $3.20{\times}10^{2}$ & $\mathrm{m^2/s^6}$ \\
EE path length           & 1.86    & 1.87    & m \\ \hline
\end{tabular}
}
\label{diffusionInferenceComparison}
\end{table}

The trained diffusion model was evaluated for its ability to infer trajectories satisfying planning constraints. To evaluate the effect of jerk regularization, inference was performed using two diffusion models: one trained with the jerk loss term and the other trained without it. The time interval between adjacent trajectory waypoints was set to $\Delta t = 40$ ms. 

Table~\ref{diffusionInferenceComparison} summarizes the performance metrics of diffusion inference over 500 test scenes. Joint jerk and EE jerk were evaluated as mean squared jerk values. The two models showed comparable mean inference times of approximately 5.7 s. The success rates were 99.6\% and 98.8\%, respectively, indicating that both models generated trajectories satisfying the planning constraints. The two models also showed similar values for manipulability and EE path length.

However, compared with the model trained without jerk regularization, the model trained with jerk regularization reduced the joint jerk and EE jerk by 36.89\% and 36.25\%, respectively. These results indicate that jerk regularization improves trajectory smoothness without degrading the overall performance of diffusion-based trajectory generation.

\subsection{Consistency Inference}

Consistency distillation was performed using the corresponding diffusion models in Section~\ref{ssDiffusionInference} as teacher models. Specifically, the consistency model with jerk regularization was distilled from the diffusion model trained with the jerk loss term, whereas the consistency model without jerk regularization was distilled from the diffusion model trained without the jerk loss term. 

Inference was performed on the same 500 test scenes, and the results are summarized in Table~\ref{consistencyInferenceComparison}. The mean inference times were 95.79 ms and 93.59 ms, respectively, reducing inference time by 98\% compared with diffusion inference. Both consistency models achieved high success rates and maintained manipulability and EE path length values comparable to those of diffusion inference.

The jerk-regularized consistency model substantially reduced joint and EE jerk compared with the model without jerk regularization. Without the jerk penalty, the joint and EE jerk increased by 264\% and 219\%, respectively, compared with the teacher diffusion model. In contrast, with jerk regularization, the joint jerk increased by 9.6\%, and the EE jerk decreased by 13.7\% relative to the teacher diffusion model. This indicates that the jerk penalty reduced the jerk increase observed in the distilled consistency model. These results indicate that consistency distillation substantially reduced inference time while maintaining high trajectory-generation performance.

\begin{table}[t!]
\centering
\caption{Consistency inference results with and without jerk regularization}
\resizebox{\columnwidth}{!}{
\begin{tabular}{lccc}
\hline
                         & w/ jerk & w/o jerk & Unit \\ \hline\hline
Inference time           & 95.79   & 93.59   & ms \\
Success rate             & 98.00   & 100.00  & \% \\
Manipulability           & 0.0931  & 0.0939  & -- \\
Joint jerk               & $6.62{\times}10^{2}$ & $3.48{\times}10^{3}$ & $\mathrm{rad^2/s^6}$ \\
EE jerk                  & $1.76{\times}10^{2}$ & $1.02{\times}10^{3}$ & $\mathrm{m^2/s^6}$ \\
EE path length           & 1.87    & 1.91    & m \\ \hline
\end{tabular}
}
\label{consistencyInferenceComparison}
\end{table}

\section{Conclusion}
\label{sdiscussionandconclusion}

In this paper, we proposed a diffusion-based motion planning framework for human-aware trajectory generation of a 6-DoF manipulator in HRI environments, accelerated by consistency distillation. When a fully specified EE pose is imposed, additional requirements such as collision avoidance and self-collision avoidance are difficult to handle as null-space secondary tasks due to limited kinematic redundancy. To address this issue, the proposed framework trains a conditional diffusion model to generate trajectory candidates conditioned on the initial joint configuration, desired EE goal pose, and obstacle representation, and applies constraint-guided sampling to reduce collisions and self-collisions. Consistency distillation was further applied to reduce the inference time required for iterative diffusion sampling, and a joint-weighted jerk regularization term was incorporated into the training loss function to improve the smoothness of the generated trajectories.

The proposed method was validated in simulation using a UR5 manipulator. The validation results demonstrate that the diffusion model generated trajectory candidates satisfying human-aware planning constraints with a high success rate. The results also demonstrate that jerk regularization improves trajectory smoothness by reducing jerk without degrading the overall trajectory-generation performance. Consistency distillation substantially reduced the number of iterative denoising steps required for diffusion inference and decreased inference time by 98\%.

Nevertheless, the present study was limited to simulation-based validation under static human configurations. It did not account for real-robot trajectory tracking errors, perception uncertainty, or dynamic human motion. Future work will validate the proposed framework on a real UR5 platform and extend it to HRI environments involving dynamic human motion and online perception.

\vfill

\end{document}